\documentclass[lettersize,journal]{IEEEtran}
\usepackage{amsmath,amsfonts,amssymb}
\usepackage{algorithmic}
\usepackage{algorithm}
\usepackage{array}
\usepackage[caption=false,font=normalsize,labelfont=sf,textfont=sf]{subfig}
\usepackage{textcomp}
\usepackage{stfloats}
\usepackage{url}
\usepackage{verbatim}
\usepackage{graphicx}
\usepackage{cite}
\usepackage[most]{tcolorbox}

\usepackage{tabularx} 
\usepackage{booktabs} 
\usepackage{graphicx} 
\usepackage{multirow} 
\usepackage{xcolor}   
\usepackage{soul}     

\usepackage{CJKutf8} 
\usepackage{multirow}
\usepackage{booktabs} 
\usepackage{microtype}
\usepackage{xcolor}
\usepackage{hyperref} 

\definecolor{LimeGreen}{rgb}{0.2, 0.8, 0.2}
\definecolor{Red}{rgb}{1, 0, 0}

\begin{document}
\begin{CJK}{UTF8}{gbsn} 

\title{Distinguishing Right from Wrong in Debates: Attribution Analysis of Chinese Harmful Memes}

\author{Weiming Wang, Junyu Lu, Han Wang, Xiaokun Zhang, Zewen Bai, Bo Xu, Liang Yang, and Hongfei Lin
\thanks{This research is supported by the Natural Science Foundation of China (No.625B2033, 62576073, 62376051), the Liaoning Provincial Natural Science Foundation Joint Fund Program (2023-MSBA-003), and the Fundamental Research Funds for the Central Universities (DUT24LAB123, DUT24MS003).}
\thanks{Weiming Wang, Junyu Lu, Zewen Bai, Bo Xu, Liang Yang, and Hongfei Lin are with the Key Laboratory of Social Computing and Cognitive Intelligence, Dalian University of Technology, Dalian, China (e-mail: 2315439390@mail.dlut.edu.cn; dut\_ljy@foxmail.com; hakuu0824@gmail.com; xubo@dlut.edu.cn; liang@dlut.edu.cn; hflin@dlut.edu.cn).}
\thanks{Han Wang is with Singapore University of Technology and Design (e-ma il: han\_wang@sutd.edu.sg).}
\thanks{Xiaokun Zhang is with the Data Intelligence Lab, Department of Computer Science, City University of Hong Kong (e-mail: dawnkun1993@gmail.com).}
}

\markboth{IEEE Transactions on Affective Computing,~Vol.~XX, No.~XX, Month~202X}%
{Wang \MakeLowercase{\textit{et al.}}: Distinguishing Right from Wrong in Debates}

\maketitle

\begin{abstract}
Research on harmful meme detection has garnered significant attention, resulting in the development of numerous datasets and methods. However, progress in detecting Chinese harmful memes lags considerably, primarily due to two challenges: first, accurately assessing a meme's harmfulness depends heavily on understanding deep cultural context; second, many memes are semantically ambiguous, making harmfulness highly subjective. To address these issues, we focus on the interpretable detection of Chinese harmful memes by constructing the first Chinese harmful meme explanation dataset, Ex-ToxiCN-MM. This dataset offers opposing interpretations, categorized as "harmful" and "non-harmful", for each meme, aiming to rigorously evaluate a model's ability to discern and comprehend ambiguous, culturally grounded content. We built a specialized knowledge base of Chinese cultural concepts and offensive vocabulary to supply models with essential prior knowledge (C-HarmKB). To address the ambiguity and lack of background knowledge in meme attribution, we have developed a comprehensive attribution analysis framework, RIKE, which includes an Attribution Knowledge Enhancement module (AKE) and a Relative Intent Reasoning module (RIR). Extensive quantitative and qualitative experiments demonstrate that our method outperforms mainstream baseline models across multiple metrics in the task of attributing harmful memes in Chinese. The code, Ex-ToxiCN-MM dataset, and Chinese Harmful Semantic Knowledge Base (C-HarmKB) involved in this study have been open-sourced at https://github.com/wimiw123/Ex-ToxiCN-MM.
\end{abstract}


\section{Introduction}
\IEEEPARstart{I}{n} recent years, internet memes have proliferated on social media platforms as a unique phenomenon of cultural dissemination. By combining images and text, these memes have become a significant medium for users to express opinions and engage socially, leveraging their humor, satirical nature, and high shareability \cite{shifman2013memes,morina2022web,zhang2025meme}. Paradoxically, their widespread popularity has also revealed a `dark side' \cite{lin2023beneath}: the rise of harmful memes. These memes often utilize subtle visual cues and double entendres to propagate hate speech, advance discriminatory ideologies, and spread misinformation, thereby posing a serious threat to both the digital ecosystem and real-world society \cite{pramanick2021momenta,brennen2021beyond,weikmann2023visual}. This challenge is particularly acute within the Chinese internet environment, where the detection of harmful memes is complicated by unique linguistic subtleties and a rich, context-dependent cultural background.

\begin{figure}[!t]
    \centering
    \includegraphics[width=0.9\columnwidth]{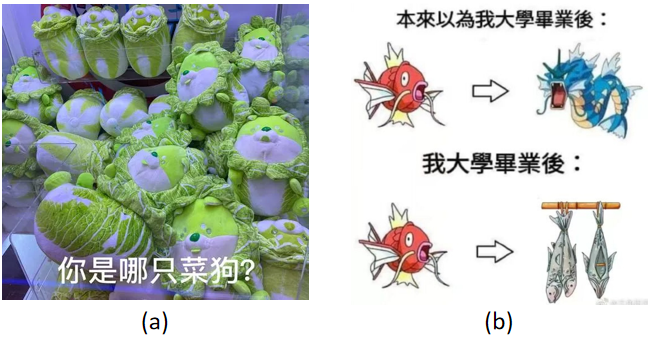} 
    \caption{Examples of harmful memes in Chinese: (a) Uses a cute ``puppy'' plushie to convey mockery. (b) Depicts the gap between idealistic expectations and harsh reality after university graduation.}
    \label{example}
\end{figure}

From the perspective of affective computing, memes are not merely combinations of images and texts but are sophisticated carriers of social affect and emotional expression \cite{DBLP:journals/taffco/SharmaSAC24}. Unlike standard multimedia content, harmful memes often exploit a mechanism of semantic incongruity—using benign or humorous visual styles to mask underlying malicious or hateful sentiments \cite{DBLP:journals/taffco/GuFFW25}. This dissonance poses a significant challenge for traditional sentiment analysis models, as they often fail to capture the subtle, high-level negative emotions (e.g., sarcasm, mockery, and covert hatred) embedded in cross-modal interactions.

Recent studies in multimodal affect recognition have highlighted the necessity of modeling these complex emotional dynamics. For instance, \cite{DBLP:journals/taffco/SharmaSAC24} emphasized that explicitly modeling emotion-enriched cues is critical for understanding meme content. Similarly, in the context of Chinese social media, \cite{DBLP:journals/taffco/GuFFW25} demonstrated that aligning visual and textual semantics is essential for detecting harmful affective content in memes. Therefore, we posit that the detection of harmful memes should be framed as a multimodal affective understanding task, where the model must not only fuse features but also decipher the implicit emotional alignment between modalities to identify harmful intent accurately.

To address the challenges posed by harmful memes, the research community has pursued several initiatives. For dataset construction, researchers have released high-quality datasets for harmful meme classification, such as HarMeme \cite{pramanick2021momenta} and MAMI \cite{gasparini2022benchmark}, significantly advancing multimodal detection techniques. More recently, a limited number of meme interpretation datasets like HatReD \cite{hee2023decoding} and ExHVV \cite{sharma2023you} have emerged, though they remain constrained by limited data volume and linguistic coverage. Initial efforts have also been made to construct datasets for Chinese harmful memes, including ToxiCN-MM \cite{lu2024towards} and MemeMind \cite{gu2025mememind}, establishing a foundation for this sub-field. However, these existing resources are often challenged by their small scale and a lack of resolution for semantic ambiguity and interpretability issues specific to the Chinese linguistic context.

However, deciphering a meme's implicit meaning requires models to not only process multimodal information but also possess deep knowledge of Chinese culture \cite{yin2025insightvision,lin2023beneath,chen2025towards}. Consider Figure \ref{example}(a), which shows a plush toy labeled with the term `vegetable dog' (cài gǒu). To understand this meme, one must recognize that `vegetable' in Chinese internet slang means `unskilled' or `low-level', not `vegetable', and that `dog' is used here as a self-deprecating suffix rather than an insult. Users without this cultural context will miss the metaphor. Consequently, a model must first be equipped with knowledge of such Chinese contextual concepts to accurately discern potential harm.

Furthermore, the detection and interpretation of harmful memes in Chinese remains underdeveloped compared to English-language contexts, primarily due to the inherent complexity of the Chinese language \cite{chakravarthi2025overview}. Determining whether a meme is `harmful' is highly subjective \cite{nguyen2024computational,sap2021annotators}, often contingent on the observer's cultural background, cognitive framework, and personal values. This subjectivity, combined with the multifaceted nature of memes, creates an ambiguous threshold for toxicity, making the judgment of harmfulness a central challenge. For example, the meme in Figure \ref{example}(b) depicts contrasting aspirations among youth after university graduation. The ambition to become a powerful `Gyarados' starkly contrasts with the Chinese internet slang `salted fish', which denotes someone who has accomplished nothing. Assessing this meme's harmfulness is inherently subjective: an optimistic individual may see it as harmless social satire, while someone experiencing depression or setbacks might feel their failures are being validated, thus perceiving it as harmful.

To advance the understanding of potentially harmful meanings in memes, we introduce a framework of conditional generation tasks. We present the Ex-ToxiCN-MM dataset, a novel Chinese meme interpretation dataset featuring explanatory annotations. Each meme is annotated with two opposing interpretations: a positive (benign) perspective and a negative (potentially harmful) one. This design supports subsequent tasks for large language models, including nuanced understanding and explanation enhancement. By providing detailed, reasoned explanations for meme content, this dataset addresses a critical gap and serves as a valuable resource for future research in Chinese meme interpretation.

To address the challenges in attributing Chinese harmful memes, we propose the attribution analysis framework RIKE, which comprises two primary models. To address the lack of contextual knowledge in interpreting Chinese memes, we adapt and enhance the Retrieval-Augmented Generation (RAG) strategy \cite{huang2024survey,ye2024r2ag,asai2024self} by constructing the Attribution Knowledge Enhancement module (AKE). This module aggregates extensive background information on Chinese cyberbullying, discriminatory language, historical events, and sociocultural contexts. It equips LLMs with the necessary prior knowledge to comprehend implicit meanings and metaphors in memes, enabling more informed interpretation generation and mitigating the challenges of semantic scarcity in Chinese. To address the subjectivity inherent in harmful meme detection, we propose the Relative Intent Reasoning module (RIR) to analyze the underlying semantics of memes, building upon the background knowledge derived from the AKE module. This approach leverages LLMs to generate and select between harmful and non-harmful interpretations, thereby performing meme classification (harmful/non-harmful) and forcing the model to conduct a fine-grained analysis of ambiguous content. Our contributions are summarized in three folds:

Our contributions are summarized in three folds:
\begin{itemize}
\item We have undertaken the first comprehensive and scientific positive-negative annotation of Chinese memes, proposing the Chinese meme interpretation dataset Ex-ToxiCN-MM.
\item We innovatively propose the relative intent reasoning framework, applying the opposing stance analysis approach of large language models to the task of Chinese meme classification and explanation generation.
\item We established a dynamic knowledge augmentation framework for harmful meme attribution tasks, constructing the first Chinese harmful semantic knowledge base, C-HarmKB, which effectively enhanced the quality of model-generated explanations.
\end{itemize}

\section{Related Work}
\subsection{Hateful Meme Detection}
The proliferation of harmful memes online has spurred significant detection efforts in recent years. For example, \cite{pramanick2021momenta} introduces the HarMeme dataset containing numerous COVID-19 memes, while the Harm-C (COVID-19) and Harm-P (US politics) datasets present multi-category attack tasks. Similarly, the MAMI dataset focuses on misogynistic memes with detailed annotations of attack types \cite{gasparini2022benchmark}. Concurrently, the detection of Chinese harmful memes has gained increasing attention. \cite{lu2024towards} proposed ToxiCN-MM, the first systematic Chinese harmful meme dataset, accompanied by the MKE baseline detector. While these contributions provide rich data resources, most remain confined to classification tasks with limited interpretability. The data annotation scheme and relative intent reasoning framework employed in this study represent a novel approach to detecting and interpreting harmful memes.

Existing research increasingly underscores the critical role of background knowledge in detecting harmful memes. For instance, \cite{kougia2023memegraphs} utilizes scene graphs to represent visual objects and their relations, combining them with textual entity recognition and background knowledge from sources like Wikidata to enhance both classification performance and interpretability. Similarly, \cite{grasso2024kermit} integrates entities identified in memes with external knowledge bases to enrich the model's contextual understanding for improved accuracy and explainability.  However, these advancements are predominantly confined to English-language contexts. Capabilities for recognizing complex cultural nuances and retrieving relevant knowledge for Chinese memes remain underdeveloped \cite{chakravarthi2025overview}. The proposed dynamic knowledge augmentation framework in this research addresses this gap.

\subsection{Large Language Models and Meme Understanding}
Large language models (LLMs) have demonstrated powerful reasoning capabilities, leading to their increasing application in harmful meme detection. For example, the HMGUARD framework employs visual art analysis guided by Chain-of-Thought (CoT) prompting to enhance model interpretation of complex meme semantics \cite{zhuang2025know}. Similarly, U-CoT+ decomposes memes into detailed textual descriptions before performing efficient zero-shot reasoning via CoT \cite{pan2025detecting}. Other studies utilize multimodal large models (MLLMs), such as LLaVA \cite{liu2023visual}, which maps visual encoder outputs as inputs to LLaMA and trains alignment layers alongside the LLM on synthetic data. Recent advancements—including sparse-sample modular LoRA fine-tuning \cite{hu2022lora}, LLM abductive reasoning combined with distillation techniques, and other state-of-the-art methods—have further broadened the scope of LLM applications in this domain. This study's novel approach combines LLM-based positive-negative interpretation analysis with retrieval-augmented generation techniques, representing an innovative attempt at interpreting harmful memes. It demonstrates superior performance across multiple baseline models.

\section{Ex-ToxiCN-MM Dataset}
To investigate the underlying logic and cultural origins of Chinese harmful memes, we introduce the first explainable Chinese harmful meme dataset, the Explanatory Toxic Chinese Multimodal Meme (Ex-ToxiCN-MM) dataset. The annotation process and statistical analysis of this dataset are detailed in the subsequent section.

\subsection{Interpretation Annotation}
For each meme, we employ a dual-opposite interpretation annotation scheme. Our core innovation lies in mandating annotators to interpret the same meme from two entirely opposing standpoints:

\textbf{Non-harmful Interpretation.} This interpretation constructs a plausible, benign perspective, simulating a superficial, culturally uninformed, or good-faith reading \cite{lin2023beneath}. Annotators are instructed to minimize potential offensive intent and interpret the meme as straightforward humor, self-deprecation, rhetorical exaggeration, or a literal reading of the text. Annotators address the question: ``Assuming the meme creator had no malicious intent, what is the most plausible non-harmful meaning?'' For the memes depicting the contrast between before and after graduation, the annotation is: ``This meme likely expresses self-deprecating humor about the gap between idealistic expectations and post-graduation reality.'' This frames the meme as a form of positive self-reflection.

\textbf{Harmful Interpretation.} This interpretation requires the annotator to assume the role of a critical analyst with deep cultural insight. The goal is to articulate the underlying harmful intent, moving beyond a surface-level reading. Annotators must leverage specific knowledge of Chinese online contexts (e.g., derogatory terms, puns, social biases, and historical references) to explain how the meme constitutes an attack, discrimination, or propagation of negative values. The annotation must provide a complete chain of reasoning, not merely a conclusion. For the ``vegetable dog'' meme, the annotation is: ``This meme uses the derogatory term `vegetable dog' (meaning `loser' or `unskilled') to insult and demean others, which can trigger conflict and reinforce negative social values.'' This explanation identifies the mechanism and nature of the social harm.

\subsection{Annotation Preparation}
Given the strong reliance of Chinese memes on local slang, metaphors, and visual rhetoric, we recruited native Mandarin speakers deeply familiar with Chinese internet culture and possessing strong analytical skills for data annotation. The team consisted of six primary annotators and three senior reviewers with extensive expertise in Chinese online culture, tasked with labeling positive and negative interpretations. We developed a detailed annotation manual and conducted comprehensive training sessions for all candidates. Since definitions can significantly influence annotation outcomes, the training focused on standardizing the definition of harmful memes, clarifying interpretation requirements, instructing on tool usage, and reviewing ethical protocols. Furthermore, annotators studied the official definitions and guidelines for categories like ``Targeted Harm'' and ``General Offense'' from the ToxiCN-MM dataset. They subsequently completed three trial annotation rounds, each involving 100 memes.

Evaluators scored the annotators' explanations based on five criteria using a five-point Likert scale:
\begin{itemize}
\item \textbf{Informativeness}: Does the explanation provide novel, valuable context?
\item \textbf{Soundness}: Is the logical reasoning clear and coherent?
\item \textbf{Cultural Relevance}: Does it accurately incorporate specific Chinese cultural knowledge?
\item \textbf{Conciseness}: Is the explanation succinct and without redundancy?
\item \textbf{Persuasiveness}: Is the argument presented convincingly?
\end{itemize}

After each round, researchers held group discussions to resolve discrepancies and calibrate annotation standards, thereby reducing subjective bias. Due to the critical importance of contextual knowledge in meme comprehension, annotators were also required to analyze images for references to public figures, landmarks, events, or artworks. They documented relevant contextual information sourced from online cultural encyclopedias into a candidate list. This candidate list was subsequently screened for inclusion in the Chinese Harmful Semantic Knowledge Base (C-HarmKB).

\begin{figure*}[htpb]
    \centering
    \includegraphics[width=0.8\linewidth]{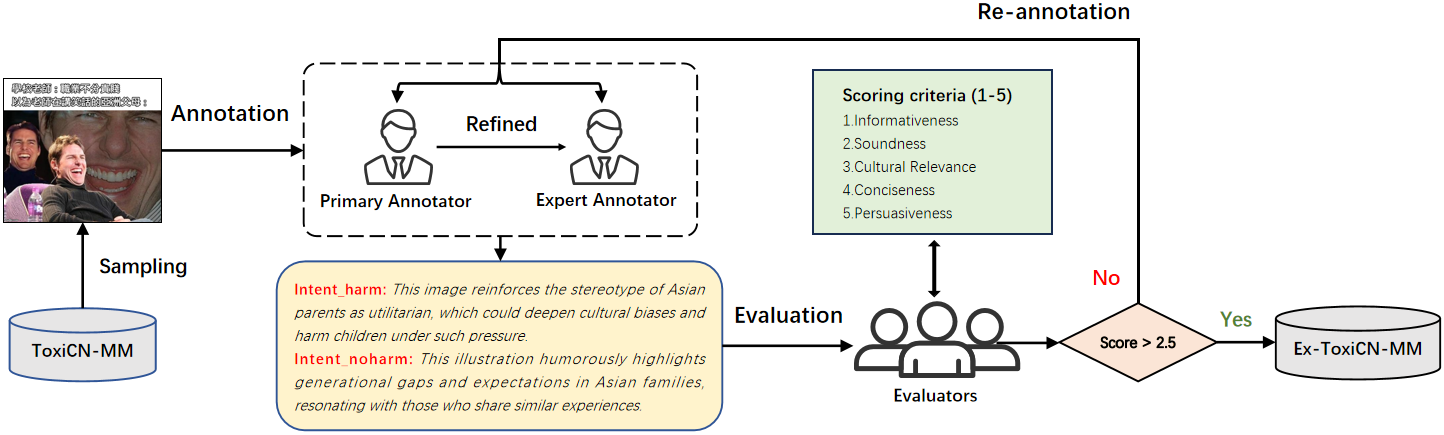}
    \caption{Annotation process for Ex-ToxiCN-MM}
    \label{annotation}
\end{figure*}

\subsection{Annotation Procedure}
To enhance linguistic diversity and ensure logical rigor in our annotations, we adapted the two-stage ``Collect-and-Edit'' framework, following the methodology of Wiegreffe and Marasovi{\'c} \cite{wiegreffe2021teach}, which has proven effective for constructing explanation datasets. The complete annotation workflow is illustrated in Figure \ref{annotation}. 
In the first stage, a primary annotator (Annotator A) generates both the non-harmful and harmful interpretations for a given meme $M$, denoted as $Exp_{n\_h}(M)$ and $Exp_{h}(M)$, respectively, aiming to produce a comprehensive initial draft.
This draft is then submitted to a second reviewer (Annotator B) for critical review and meticulous editing. The key objectives of this review are to:
\begin{itemize}
\item Verify the accuracy of cultural references, internet slang, and other knowledge cited in the interpretations.
\item Strengthen the logical coherence and persuasiveness of the explanations.
\item Correct grammatical errors and improve textual fluency.
\item If Annotator B disagrees with the initial interpretation or proposes a superior alternative, the two annotators discuss to reach a consensus. Cases without consensus are escalated to a researcher for final adjudication.
\end{itemize}

Following the annotation, we conducted a quality assessment to quantify the annotation quality. Three reviewers evaluated each explanation using a five-point Likert scale across five dimensions: Informativeness, Soundness, Cultural Relevance, Conciseness, and Persuasiveness \cite{chen2025multimodal,sullivan2013analyzing}. To control for the influence of length, annotators were instructed to target explanations of approximately 30 words. All samples where any reviewer scored below 2.5 on any dimension were flagged.

\begin{table}[!t]
    \centering
    \caption{Evaluation results on datasets with different Explanations. Five evaluation criteria: Informativeness (Info.), Soundness (Sound.), Cultural Relevance (Cult.), Conciseness (Conc.), Persuasiveness (Pers.).}
    \label{tab:eval1}
    \small 
    \begin{tabular}{@{}l l c c c c c@{}}
    \toprule
    Dataset & Explanation & \multicolumn{5}{c}{Metrics} \\
    & & Info. & Sound. & Cult. & Conc. & Pers. \\
    \midrule
    \multirow{2}{*}{Harmful} & Harmful & 4.72 & 4.81 & 4.05 & 4.53 & 4.79 \\
     & non-harmful & 3.85 & 4.25 & 3.75 & 4.61 & 3.92 \\
    \multirow{2}{*}{non-harmful} & Harmful & 3.53 & 3.68 & 3.49 & 4.15 & 3.27 \\
     & non-harmful & 4.21 & 4.65 & 3.55 & 4.58 & 4.71 \\
    \bottomrule
    \end{tabular}
\end{table}

To mitigate bias, a separate group of annotators, who had not previously encountered these samples, re-annotated them. This process was repeated until all scores met the quality standard.

\subsection{Statistics Description}
The final Ex-ToxiCN-MM dataset contains 7,042 samples, comprising 3,735 harmful and 3,307 non-harmful memes. The harmful memes are categorized into four themes—Targeted Harm, Sexual Innuendo, General Offense, and Disparaging Culture—with the distribution detailed in Table \ref{tab:data}. General Offense represents the largest category, suggesting that content attacking group identities (e.g., based on region, gender, or nationality) is a primary source of toxicity in Chinese online meme culture. A quantitative analysis of the 14,084 explanatory texts shows an average length of 33.29 characters with a standard deviation of 7.40. As shown in Table \ref{tab:eval1}, the dataset achieved high scores across the five evaluation metrics. Specifically, the harmful explanations attained high ratings in Persuasiveness, Informativeness, and Cultural Relevance, demonstrating that our annotation protocol successfully produces concise yet culturally-grounded explanations.

\begin{table}[!t]
  \centering
  \caption{Ex-ToxiCN-MM dataset statistics. The table shows sample counts for training and testing datasets across different harmful content categories: target harmful content (Tg.), general offense (Off.), sexual innuendo (Sex.), and depressing culture (Disp.).}
  \label{tab:data}
  \small
  \begin{tabular}{l c c c c c}
  \toprule
  \textbf{Split} & \textbf{Harmful} & \multicolumn{4}{c}{\textbf{Harmful Type}} \\
  & & \textbf{Tg.} & \textbf{Off.} & \textbf{Sex.} & \textbf{Disp.} \\
  \midrule
  Train & 2,988 & 711 & 862 & 1,093 & 862 \\
  Test & 747 & 178 & 216 & 274 & 178 \\
  \midrule
  Total & 3,735 & 889 & 1,078 & 1,367 & 1,040 \\
  \bottomrule
  \end{tabular}
\end{table}

\section{Methodology}
In the following sections, we will elaborate on RIKE, an attribution framework for Chinese harmful memes. This framework addresses the lack of contextual knowledge and subjective challenges encountered in the attribution analysis of Chinese harmful memes.

\begin{figure*}[htpb]
    \centering
    \includegraphics[width=0.9\linewidth]{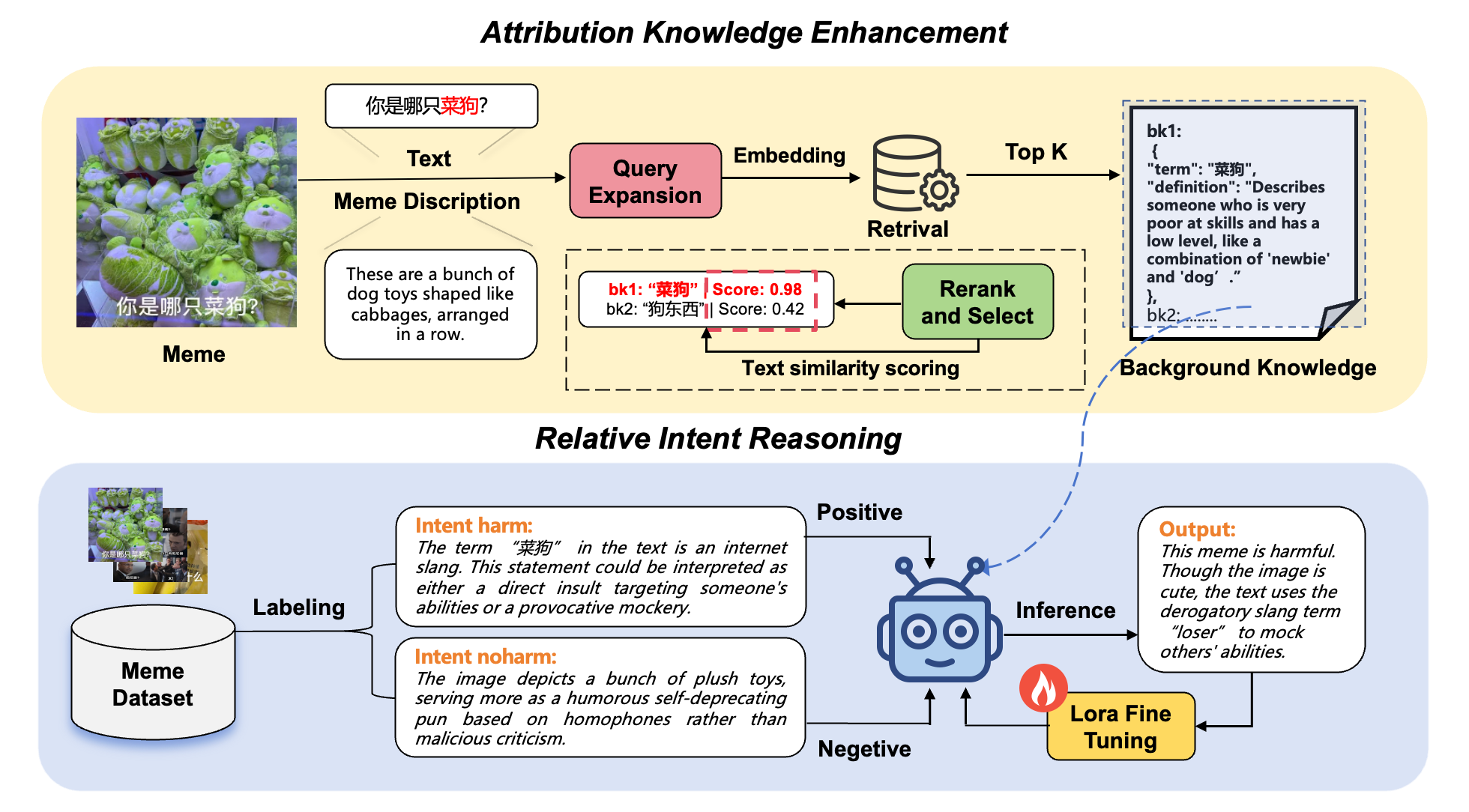}
    \caption{Illustration of the Attribution Analysis Framework, integrating the Relative Intent Reasoning module and Attribution Knowledge Enhancement module.}
    \label{intro}
\end{figure*}

\subsection{Overview}
The analysis of harmful memes remains challenging for many models due to their lack of contextual knowledge and the inherent ambiguity of memes. To address these limitations, our framework (RIKE) processes memes in two sequential stages: (1) Attribution Knowledge Enhancement (AKE) and (2) Relative Intent Reasoning (RIR), as shown in Figure \ref{intro}. In the first stage, we employ a Retrieval-Augmented Generation (RAG) mechanism to construct a dynamic knowledge enhancement framework for analyzing new memes. 
In the second stage, we fine-tune the LLM on paired harmful and non-harmful interpretations of each meme, equipping it to reason over competing perspectives.
This allows the model to retrieve the most relevant contextual information, thereby generating concise, reasonable, and culturally appropriate explanations for the toxicity of Chinese memes.

\subsection{Attribution Knowledge Enhancement}
The semantic complexity of Chinese often causes mainstream language generation models to lack the necessary cultural context and semantic depth when interpreting harmful memes. To overcome this limitation, we propose a dynamic knowledge augmentation framework (AKE), which comprises knowledge base construction, dynamic querying, and multi-stage retrieval.

\subsubsection{Knowledge Base Construction}
To mitigate the limited understanding of Chinese slang and cultural context in existing models, we constructed the first Chinese Harmful Semantic Knowledge Base (C-HarmKB), comprising 2,870 entries. This resource was built by expanding upon an existing Chinese insult lexicon. We harvested relevant terms and their definitions from encyclopedic sources, including Baidu Baike and Wikipedia, as well as specialized subculture forums, using automated web crawlers. The knowledge base was further enriched with contextual information gathered during the creation of the Ex-ToxiCN-MM dataset. All candidate terms underwent a rigorous, multi-step annotation process: three native Chinese speakers with expertise in internet culture categorized each term into predefined classes (e.g., ``Sexism,'' ``Racism,'' ``Region,'' ``LGBTQ,'' ``Others'') and authored detailed definitions elucidating their cultural context, origin, and harmful connotations.

\subsubsection{RAG Pipeline Design}
During system initialization, the textual content of the knowledge base is segmented and converted into vector representations using a pre-trained Chinese word embedding model. These representations, which encode the vocabulary, classifications, and definitions, are stored in a vector database.

\textbf{Dynamic Query Expansion:} The textual content of memes is often too concise or obscure for direct retrieval to be effective. To address this, we introduce a dynamic query expansion module. This module employs a lightweight Large Language Model (LLM) to analyze the meme's text $T$ and image description $X$ guided by specific prompt instructions, the LLM generates a set of relevant keywords and concepts $T_{set}$. This expanded set is then merged with the original text and image description to form a comprehensive query set $Q_{set}$, defined as follows:

\begin{small}
\begin{equation}
Q_{\text{set}} = \{q|q \in (T \cup X \cup \{T_{\text{set}}\}) \cap q \neq \emptyset\}
\end{equation}
\end{small}

\textbf{Hybrid Retrieval and Rerank:} To simultaneously leverage the precision of keyword matching and the semantic generalization capabilities of vector retrieval \cite{lu2022hyrr,rayo2025hybrid}, we constructed a hybrid retriever combining BM25 keyword retrieval with dense vector retrieval. For any document $d$ and query $q$, its hybrid relevance score $S_{hybrid}$ can be expressed as:

\begin{small}
\begin{equation}
S_{\text{hybrid}}(d,q) = w_{\text{bm25}} \cdot S_{\text{BM25}}(d,q) + w_{\text{dense}} \cdot S_{\text{dense}}(d,q)
\end{equation}
\end{small}

Simultaneously, we employ a reordering model where, for each document-query pair (d, q), the reordering component constructs a prompt and forces a ``yes'' or ``no'' response. The model calculates logits for the labels at the last token position, and the document relevance score is defined as the posterior probability of the ``yes'' token:

\begin{small}
\begin{equation}
P_{\text{rel}}(d|q) =\frac{\exp(\text{logits}_{\text{yes}})}{\exp(\text{logits}_{\text{yes}}) + \exp(\text{logits}_{\text{no}})}
\end{equation}
\end{small}

Subsequently, we introduce a gatekeeping threshold $\tau_{\text{rel}}$ to filter out background information irrelevant to meme interpretation or of low quality. Only $P_{\text{rel}}(d|q) \geq \tau_{\text{rel}}$ then is the document retained. After acquiring the most relevant background knowledge, we integrate this knowledge into model fine-tuning and reasoning instructions to generate explanations. The purpose of this knowledge-enhanced framework is to ensure generated explanations are both accurate and interpretable—covering Chinese cultural and online contexts while suppressing the negative impact of noisy background knowledge, thereby enhancing the comprehensibility of harmful meme interpretations.

\subsection{Relative Intent Reasoning}

To mitigate the subjectivity inherent in Chinese harmful meme detection, we propose a Relative Intent Reasoning (RIR) module. RIR trains the model to contrast two competing interpretive trajectories, one supporting a harmful label and the other supporting a non-harmful label, while being grounded in meme background knowledge retrieved by the Attribution Knowledge Enhancement (AKE) module.

We represent a meme as a tuple: $M = (I, T, X)$,
where $I$ denotes the image, $T$ is the embedded text, and $X$ is a textual description of the image.
Given $M$, AKE retrieves the top-$k$ relevant knowledge fragments, denoted as $K(M)$, which provide cultural and contextual cues that help the model move beyond surface-level semantics.

To explicitly model interpretive subjectivity, we introduce two opposing perspectives for each meme in the Ex-ToxiCN-MM dataset, e.g., $Exp_{n\_h}(M)$ and $Exp_{h}(M)$.
RIR formulates the decision process as a debate-style comparison between these two interpretive paths, requiring the model to determine which perspective is more plausible in light of the meme content and the retrieved background knowledge.
This helps the model better characterize the ambiguous decision boundary between harmful and non-harmful cases.

To fully leverage the guiding effect of these relative interpretive paths, we further train the model via instruction tuning. Specifically, we optimize the model on two complementary tasks: \textit{Harmful Meme Detection} and \textit{Decision Explanation Generation}.
In the detection setting, the model predicts the final decision label $y^*$ conditioned on the meme content, integrating two paths and the AKE knowledge:
\begin{equation}
\begin{small}
y^* = C_{\theta}(I, T, X, Exp_{n\_h}(M), Exp_{h}(M), K(M)).
\end{small}
\end{equation}
Given annotated labels (harmful / non-harmful), we minimize the classification loss:

\begin{small}
\begin{equation}
L_{\text{clf}} = -\log P(y^*|\text{M}, Exp_{n\_h}(M), Exp_{h}(M), K(M))
\end{equation}
\end{small}

Beyond classification, we further train the model to generate faithful decision explanations, thereby improving its ability to organize relevant evidence while integrating the two interpretive paths. Let $\mathrm{Exp}(M)$ denote the model-generated explanation. Conditioned on the same inputs, we optimize:

\begin{equation}
\begin{small}
L_{\text{itf}} = -\log P(Exp(M)|M, Exp_{n\_h}(M), Exp_{h}(M), K(M))
\end{small}
\end{equation}

Through this joint training procedure, the model is encouraged to make decisions by explicitly contrasting the two perspectives and grounding its judgment in the retrieved background knowledge, thereby improving both classification robustness and explanation quality.

In the implementation of the Relative Intent Reasoning (RIR) phase, we adopt the Low-Rank Adaptation technique (LoRA) to efficiently fine-tune $C_{\theta}$.
The specific instruction template is shown as follows:

\begin{tcolorbox}[title=Relative Intent Reasoning Prompt]
\begin{small}

You are a meme analysis expert proficient in Chinese internet culture and social psychology.
Given a Chinese meme and two opposing interpretive perspectives (harmful vs. non-harmful), determine whether the meme expresses harmful content. Provide a clear decision and a concise justification. \\

\textbf{Input Information:}\\
Meme Image: \textit{I},\\
Meme Text: \textit{T},\\
Meme Image Description: \textit{X},\\
Meme Background Knowledge: \textit{K(M)},\\
Interpretation of Non-harmful Perspective: \textit{{$\mathrm{Exp_{n\_h}}(M)$}},\\
Interpretation of Harmful Perspective: \textit{{$\mathrm{Exp_{h}}(M)$}}\\

\textbf{Output Requirements:}\\
Answer: \{Harmful / Non-harmful\},\\
Reason: \textit{{$\mathrm{Exp}(M)$}}
\end{small}
\end{tcolorbox}

\section{Experiment}

\subsection{Experimental Setup}
\textbf{Dataset and Tasks.} We evaluate our model on the proposed Ex-ToxiCN-MM dataset, formulating harmful meme attribution analysis through two core tasks: harmful meme detection and harmful explanation generation. The harmful meme detection task assesses the model's ability to discriminate between harmful and non-harmful memes across the entire test set. 
For samples classified as harmful, the explanation generation task requires the model to produce a natural language rationale clarifying its decision. To ensure a fair evaluation of explanatory capability, this second task is performed exclusively on the harmful subset of the test set. This design isolates explanation quality from classification bias, directly evaluating the model's capacity to attribute harmful semantics.

\textbf{Evaluation Metrics.} The model is evaluated on the test set using multi-dimensional metrics. To comprehensively assess the classifier's robustness against class imbalance, we report its accuracy alongside the precision, recall, and F1-score specifically for the ``harmful'' class. The quality of the generated explanations is evaluated through both automated metrics and human-aligned assessment. For automated evaluation, we employ ROUGE and BLEU to measure the n-gram overlap and fluency against human-annotated references \cite{ganesan2018rouge,evtikhiev2023out}. Furthermore, we utilize the powerful closed-source multimodal model, Qwen-VL-Plus, as an automated judge \cite{gu2024survey}. It scores the explanations on a 5-point Likert scale across five dimensions: informativeness, soundness, cultural relevance, conciseness, and persuasiveness.

\subsection{Baseline}
In our experiments on classification and explanation generation, we evaluated a diverse set of open- and closed-source models. As the primary baseline, we used the Qwen2.5-VL series, a Chinese multimodal model with strong visual recognition, localization, text understanding, and cross-modal reasoning capabilities \cite{bai2025qwen2}. We examined its scaling behavior on harmful meme attribution across the 3B and 8B variants. For architectural diversity, we included InternVL3\_5-8B, a dialogue-oriented multimodal model with strong visual reasoning, serving as a cross-architecture control \cite{chen2024internvl}. We also evaluated LLaVA-1.5-7B, a widely adopted earlier vision–language model with a reusable architecture \cite{liu2024improved}. In addition, we tested two closed-source models, Qwen-VL-Max and GLM-4V-Plus.

\subsection{Assessing Changeability}

\begin{table*}[!t]
  \centering
  \caption{Performance comparison of different models and training strategies on harmful meme detection. RIR: Relative Intent Reasoning, BK: Background Knowledge, Base: Direct Reasoning, SFT: Supervised Fine-tuning, RIKE: Complete Attribution Framework, where the closed-source models are used without SFT.}
  \label{tab:performance_comparison}
  \resizebox{\textwidth}{!}{%
  \begin{tabular}{l cccc c cccc c cccc}
  \toprule
   & \multicolumn{4}{c}{\textbf{Qwen2.5-VL-3B}} & & \multicolumn{4}{c}{\textbf{Qwen2.5-VL-7B}} & & \multicolumn{4}{c}{\textbf{Qwen-VL-max}} \\
  \cmidrule{2-5} \cmidrule{7-10} \cmidrule{12-15}
   & Acc. & P & R & F1 & & Acc. & P & R & F1 & & Acc. & P & R & F1 \\
  \midrule
  Base    & 0.569 & 0.673 & 0.364 & 0.472 & & 0.583 & 0.912 & 0.236 & 0.375 & & 0.719 & \textbf{0.875} & 0.546 & 0.673 \\
  RIR     & 0.566 & 0.567 & 0.772 & 0.654 & & 0.643 & 0.805 & 0.431 & 0.562 & & 0.731 & 0.780 & 0.686 & 0.730 \\
  RIR+BK  & 0.619 & 0.595 & 0.878 & 0.710 & & 0.702 & 0.834 & 0.546 & 0.660 & & - & - & - & - \\
  SFT     & 0.826 & \textbf{0.945} & 0.714 & 0.813 & & 0.869 & 0.953 & 0.793 & 0.866 & & - & - & - & - \\
  SFT+RIR & 0.943 & 0.937 & 0.956 & 0.946 & & 0.956 & \textbf{0.962} & 0.955 & 0.958 & & - & - & - & - \\
  RIKE    & \textbf{0.948} & 0.940 & \textbf{0.964} & \textbf{0.952} & & \textbf{0.957} & 0.949 & \textbf{0.971} & \textbf{0.960} & & \textbf{0.757} & 0.784 & \textbf{0.746} & \textbf{0.765} \\
  
  \midrule
  \addlinespace[0.5em] 
  
   & \multicolumn{4}{c}{\textbf{InternVL3\_5-8B}} & & \multicolumn{4}{c}{\textbf{LLaVa-1.5-7b}} & & \multicolumn{4}{c}{\textbf{GlM-4v-flash}} \\
  \cmidrule{2-5} \cmidrule{7-10} \cmidrule{12-15}
   & Acc. & P & R & F1 & & Acc. & P & R & F1 & & Acc. & P & R & F2 \\
  \midrule
  Base    & 0.596 & 0.837 & 0.296 & 0.437 & & 0.509 & 0.572 & 0.292 & 0.387 & & 0.608 & \textbf{0.843} & 0.283 & 0.423 \\
  RIR     & 0.614 & 0.810 & 0.355 & 0.494 & & 0.557 & 0.583 & 0.580 & 0.581 & & 0.649 & 0.747 & 0.470 & 0.577 \\
  RIR+BK  & 0.674 & 0.840 & 0.477 & 0.608 & & 0.592 & 0.607 & 0.655 & 0.630 & & - & - & - & - \\
  SFT     & 0.866 & 0.964 & 0.776 & 0.860 & & 0.723 & \textbf{0.986} & 0.485 & 0.650 & & - & - & - & - \\
  SFT+RIR & 0.966 & \textbf{0.974} & 0.961 & 0.968 & & 0.897 & 0.969 & 0.833 & 0.896 & & - & - & - & - \\
  RIKE    & \textbf{0.967} & 0.969 & \textbf{0.968} & \textbf{0.969} & & \textbf{0.915} & 0.969 & \textbf{0.868} & \textbf{0.915} & & \textbf{0.705} & 0.727 & \textbf{0.674} & \textbf{0.700} \\
  \bottomrule
  \end{tabular}%
  }
\end{table*}

The result of harmful meme detection is shown in Table \ref{tab:performance_comparison}.
Based on the results, we can draw the following conclusions:

(1) We observe that under zero-shot settings, open-source models exhibit relatively weak classification performance, with accuracy generally below 60\%. In contrast, more powerful closed-source models achieve substantially stronger overall results, highlighting the critical role of model capacity and large-scale pretraining in this task.
After introducing instruction fine-tuning, the performance of open-source models improves markedly, indicating successful acquisition of task-specific reasoning patterns. 

(2) In cross-architecture comparisons at comparable parameter scales (e.g., 7B and 8B), Qwen2.5-VL shows only a modest performance gap relative to InternVL3.5. By contrast, LLaVA-1.5 consistently underperforms on this task, likely due to its predominantly English-oriented pretraining and less effective alignment for Chinese multimodal inputs. Overall, these results underscore that Chinese harmful meme analysis, as an inherently context-sensitive task, relies on both deep Chinese linguistic/cultural understanding and robust vision–language alignment.
Within the Qwen2.5-VL family (3B vs. 7B), results indicate that the larger variant generally performs better across most experimental settings, suggesting that scaling up model parameters can yield performance gains for this task.

(3) We further demonstrate the effectiveness of each module through ablation studies. Incorporating Relative Intent Reasoning (RIR) substantially enhanced the model’s ability to detect harmful content. For example, the recall of harmful memes for Qwen2.5-VL-3B increased from 0.364 to 0.772 after introducing RIR, accompanied by a clear improvement in the F1 score. This indicates that explicitly encouraging the model to reason over opposing viewpoints helps reduce false negatives when handling semantically ambiguous memes.
Further integrating the knowledge augmentation module (AKE) led to consistent gains in both accuracy and recall. Specifically, Qwen2.5-VL-3B achieved an approximately 5.3\% increase in accuracy and over a 5.6\% improvement in F1 when RIR was combined with background knowledge (BK). These results suggest that injecting external knowledge effectively compensates for the model’s limitations in cultural understanding, thereby mitigating misjudgments caused by semantic gaps.

\begin{table*}[!t]
  \centering
  \caption{Evaluation results of different Models on multiple metrics. Five evaluation criteria: Informativeness (Info.), Soundness (Sound.), Cultural Relevance (Cult.), Conciseness (Conc.), Persuasiveness (Pers.). RIKE for the complete attribution framework, where the closed-source models are used without SFT.}
  \label{tab:main_results}
  \small
  \begin{tabular}{
    >{\raggedright\arraybackslash}p{2.2cm} 
    >{\raggedright\arraybackslash}p{1.5cm} 
    >{\centering\arraybackslash}p{0.9cm} 
    >{\centering\arraybackslash}p{0.9cm} 
    >{\centering\arraybackslash}p{0.9cm} 
    >{\centering\arraybackslash}p{0.9cm} 
    >{\centering\arraybackslash}p{0.9cm} 
    >{\centering\arraybackslash}p{1.2cm} 
    >{\centering\arraybackslash}p{1.2cm}
  }
  \toprule
  \textbf{Module} & \textbf{Method} & \multicolumn{5}{c}{\textbf{Module Evaluation}} & \multicolumn{2}{c}{\textbf{Automatic Metrics}} \\
  & & Info. & Sound. & Cult. & Conc. & Pers. & BLEU-4 & ROUGE-L \\
  \midrule
  \multirow{2}{*}{Qwen2.5-VL-7B} 
    & Base & 2.73 & 3.64 & 2.51 & 4.50 & 2.97 & 14.17 & 29.31 \\
    & RIKE & 2.84 & 3.76 & 2.64 & 4.57 & 3.15 & \underline{51.69} & \underline{62.51} \\
  \addlinespace[0.1em]
  
  \multirow{2}{*}{InternVL3\_5-8B}
    & Base & 2.75 & 3.70 & 2.63 & \textbf{4.79} & 3.14 & 8.86 & 28.28 \\
    & RIKE & 2.90 & 3.83 & 2.71 & \underline{4.63} & 3.22 & \textbf{53.48} & \textbf{64.38} \\
  \addlinespace[0.1em]
  
  \multirow{2}{*}{LLaVa-1.5-7b}
    & Base & 2.52 & 3.32 & 2.58 & 3.77 & 2.64 & 11.20 & 25.64 \\
    & RIKE & 2.57 & 3.48 & 2.36 & 4.44 & 2.75 & 51.20 & 44.25 \\
  \addlinespace[0.1em]
  
  \multirow{2}{*}{Qwen-VL-Max}
    & Base & 3.05 & 3.83 & \underline{3.06} & 4.34 & 3.31 & 9.94 & 1.09 \\
    & RIKE & \textbf{3.11} & \underline{3.90} & \textbf{3.13} & 4.45 & \textbf{3.39} & 10.17 & 1.01 \\
  \addlinespace[0.1em]
  
  \multirow{2}{*}{GLM-4V-Flash}
    & Base & 3.04 & 3.88 & 3.01 & 3.81 & 3.26 & 2.94 & 18.98 \\
    & RIKE & \underline{3.08} & \textbf{3.98} & 2.87 & 4.41 & \underline{3.37} & 9.48 & 32.30 \\
  \addlinespace[0.1em]
  
  Ex-ToxiCN-MM
    & human & 3.14 & 4.03 & 2.90 & 4.79 & 3.46 & - & - \\
  \bottomrule
  \end{tabular}
\end{table*}

\subsection{Assessing Generation Quality}
For explanation generation, models were evaluated under the comprehensive framework (RIKE) combining oppositional reasoning (RIR) and knowledge augmentation (AKE), as shown in Table \ref{tab:main_results}. In automatic evaluations, open-source models showed markedly improved surface-level alignment with reference explanations after integrating the framework. Qwen2.5-VL-7B's BLEU-4 score surged from 14.17 to 51.69, and its ROUGE-L from 29.31 to 62.51. Similarly, the fine-tuned InternVL3\_5-8B and llava-1.5-7b models saw their BLEU-4 scores rise from 8.86 and 11.20 to 53.48 and 51.20, respectively. In contrast, closed-source models like Qwen-VL-Max showed limited gains on these automated metrics, likely because their inherent generation style differs from the reference texts. This finding underscores the need for more nuanced evaluation methods beyond n-gram matching for high-performing models.

To assess the substantive quality of explanations, we employed Qwen3-VL-Plus as an evaluator. The results confirm that the full framework delivers comprehensive improvements across all tested models. For Qwen2.5-VL-7B, scores increased in informativeness (to 2.84), soundness (to 3.76), cultural relevance (to 2.64), and persuasiveness (to 3.15). While closed-source models already demonstrated strong performance in baseline settings, the framework provided further gains; Qwen-VL-Max, for example, saw its scores for informativeness, soundness, and cultural relevance rise from 3.05, 3.83, and 3.06 to 3.11, 3.90, and 3.13, respectively. This demonstrates that the framework provides effective guidance even for high-performance models, enhancing their explanation quality. The experimental results validate that the relative intent reasoning and knowledge augmentation framework effectively addresses the core challenges of Chinese harmful memes' semantic ambiguity and contextual knowledge gaps—achieving significant performance gains in both classification and explanation generation tasks.

\subsection{Case Studies}
\begin{figure*}[t] 
  \centering
  \includegraphics[width=\textwidth]{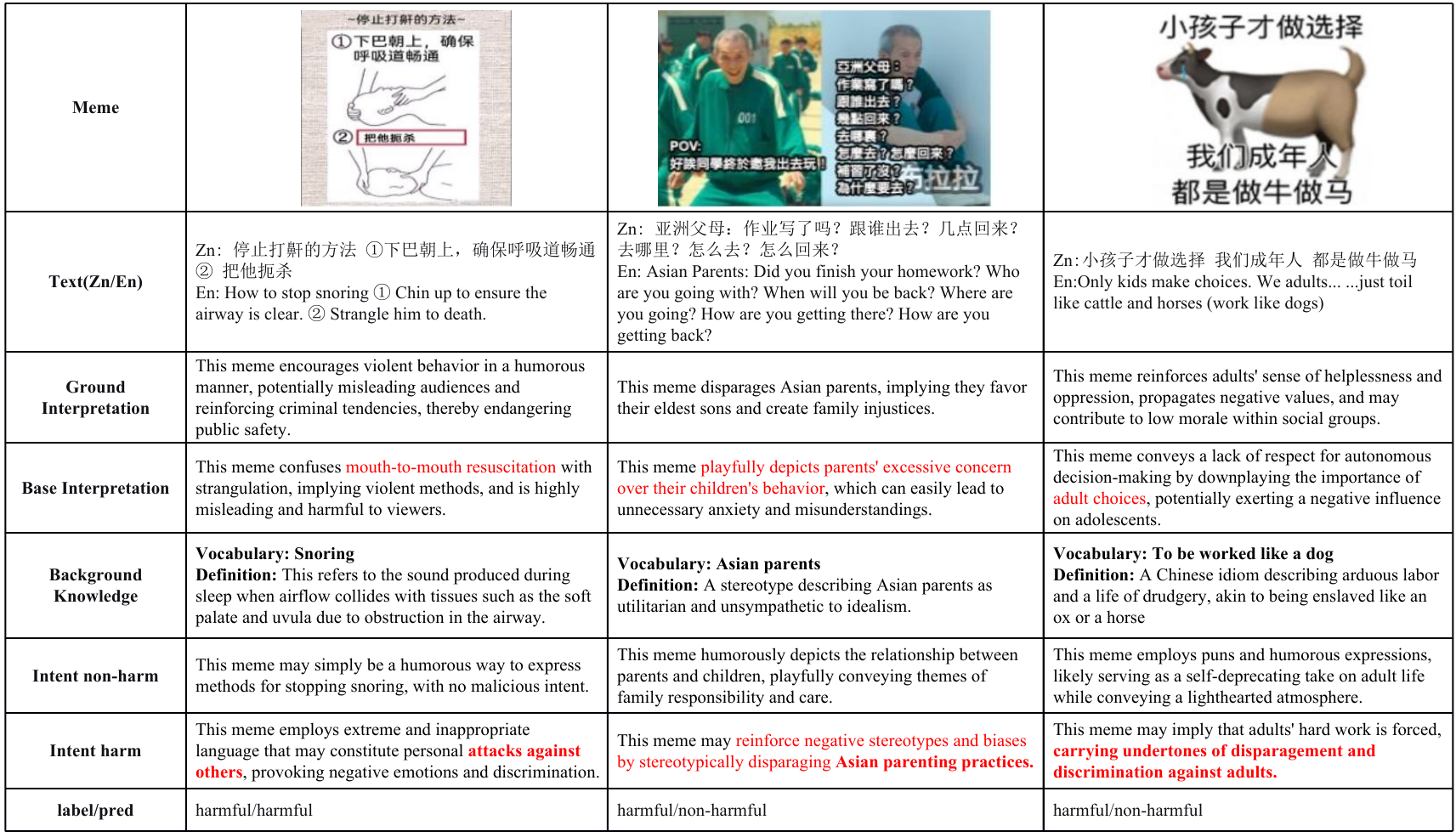}
  \caption{Case study of attribution process. Presents three types of interpretive and retrieval knowledge derived from annotated datasets, baseline models, and framework outputs.}
  \label{case study}
\end{figure*}

To further illustrate the effectiveness of the proposed attribution framework, we provide case studies in Figure \ref{case study}. In all three cases, the contextual knowledge (Background Knowledge) retrieved by the knowledge-enhanced component addressed the baseline model's deficiencies in understanding cultural context. For example, in case(a), the baseline model (Base Interpretation) without knowledge enhancement interpreted the phrase as emphasizing the importance of adult choices. In contrast, the enhanced model accurately identified the literal meaning of ``working like a horse'' as referring to labor intensity, yielding a more reasonable explanation (Intent harm). Similarly, in case(b), the knowledge-augmented model successfully captured the satirical stereotype associated with ``Asian parents.'' The relative intent reasoning framework further enhanced the explanations' depth and persuasiveness. By compelling the model to reason from opposing perspectives (Intent non-harm), it avoided oversimplified misjudgments and produced final interpretations that were more logical, critical, and aligned with human annotations. For instance, in case(c), the baseline model generated a hallucination, misinterpreting the meme's action as ``mouth-to-mouth resuscitation''. By considering the potential for a non-harmful interpretation, the model instead focused on the distinct expression of ``stopping snoring'', leading to the correct inference of an extreme physical assault (strangulation) rather than a rescue effort.

Although the model correctly classified the overtly violent content in case(a) as \textit{harmful}, it produced false negative predictions (\textit{harmless}) for case(b) and case(c).
In case(b), despite recognizing the “Asian parents” stereotype, the model may have overemphasized the generated harmless perspective (family care) while failing to recognize psychological pressure as a toxic manifestation.
Similarly, in case(c), the model interpreted self-deprecating humor (“working like a beast of burden”) as a harmless joke, overlooking the negative values and despair it propagates.
These cases illustrate ongoing challenges in detecting harmful memes in Chinese: distinguishing benign humor from harmful satire requires not only cultural awareness but also subjective, nuanced calibration of social values.
Future research may thus explore richer end-to-end multimodal architectures to more precisely capture semantic subtleties and achieve value alignment.

\section{Conclusion}
This study addresses two major challenges in detecting Chinese harmful memes: strong cultural dependency and semantic ambiguity. To this end, we systematically construct the first Chinese harmful meme explanation dataset, Ex-ToxiCN-MM, which features opposing interpretations. We propose an attribution analysis framework that integrates LoRA fine-tuning with the first Chinese harmful semantic knowledge base, C-HarmKB, to fuse relative intent reasoning with knowledge augmentation. This framework guides the model to perform in-depth analysis within ambiguous contexts while providing essential cultural context to support its decision-making. Experimental results demonstrate that our approach enhances the accuracy and interpretability of harmful meme attribution analysis in both classification and explanation generation tasks. This work establishes a foundational dataset and a robust methodological benchmark for future research.


\bibliographystyle{IEEEtran} 
\bibliography{mutihate}

\newpage

\end{CJK}

\end{document}